\documentclass{article}
\usepackage{spconf,amsmath,graphicx}
\usepackage{subfigure}

\usepackage{pgfplots}

\usepackage{color} 

\title{CNN-aware Binary Map for General Semantic Segmentation}
\name{Mahdyar Ravanbakhsh$^{\star 1}$, Hossein Mousavi\sthanks{Equal contribution.}$^{ 2}$,  Moin Nabi $^{ 3}$, Mohammad Rastegari $^{ 4}$, Carlo Regazzoni $^{1}$}

\address{$^{1}$ University of Genova  $^{2}$  Istituto Italiano di Tecnologia
  $^{3}$ University of Trento  $^{4}$  Allen Institute for AI} 
\begin{document}
%
\maketitle
%

\begin{abstract}
In this paper we introduce a novel method for general semantic segmentation that can benefit from general semantics of Convolutional Neural Network (CNN). Our segmentation proposes visually and semantically coherent image segments. We use binary encoding of CNN features to overcome the difficulty of the clustering on the high-dimensional CNN feature space. These binary codes are very robust against noise and non-semantic changes in the image. These binary encoding can be embedded into the CNN as an extra layer at the end of the network. This results in real-time segmentation. To the best of our knowledge our method is the first attempt on general semantic image segmentation using CNN. All the previous papers were limited to few number of category of the images (e.g. PASCAL VOC). Experiments show that our segmentation algorithm outperform the state-of-the-art non-semantic segmentation methods by large margin.
\end{abstract}
%


\section{Introduction}
\label{sec:intro}
Image segmentation is a challenging task in computer vision that can specify the visual elements in an image. These elements can be used as the building blocks for any image understanding method. Traditionally, these image segments are optimized to be semantic (e.g. be an object, part of an object, or part of a scene) and visually coherent; This means that nearby pixels in each segment must have similar intensity \cite{felzenszwalb2004efficient,achanta2012slic,khoreva2016improved}. Semantic image segmentation has been proposed in several articles \cite{girshick2014rich,long2015fully,chen2014semantic,zheng2015conditional,noh2015learning}. All of these methods are limited to a narrow scope of semantics. They can only find the segments belong to \emph{few categories} of objects (e.g. 20 categories in PASCAL VOC dataset). In this paper a method is proposed that can find \emph{general} semantic segments.         

\begin{figure}[htb]
\centering
\centerline{\includegraphics[width=0.47\textwidth]{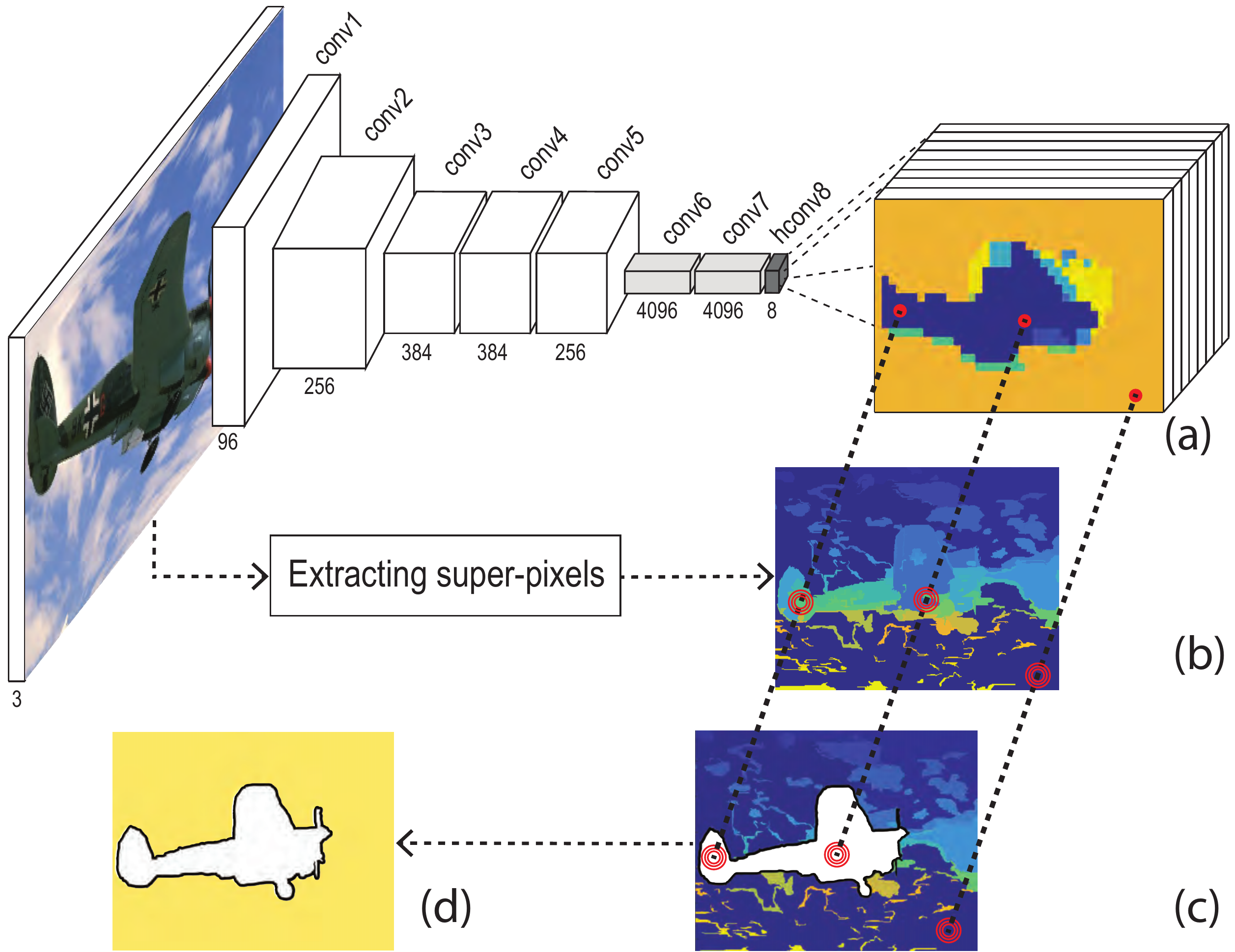}}
\caption{\small Method overview: given an image the semantic binary map is extracted by forward-pass of image through the net (a), a low-level superpixel is extracted (b), the binary code of each superpixel is assigned using the corresponding region on the binary map (c), finally semantic segmentation is generated merging superpixels with similar binary patterns (d).}
\label{fig:overview}
\end{figure}

Recently there has been a remarkable progress in computer vision through Deep Neural Networks. More specifically, with Convolutional Neural Networks (CNNs) an end-to-end object recognition has been created \cite{alexnet, simonyan2014very, szegedy2015going} that outperformed all of the previous recognition systems. These learning methods recently became more popular than traditional statistical learning techniques. CNN is a multi-layer neural network; in each layer the weights are in the shape of filters. The output of each layer is the result of convolution of filters on that layer with the input. After several layers of convolution the output can be used as a feature representation of an image. For example, in  AlexNet \cite{alexnet} architecture the output of \emph{fc7} layer has been used extensively as a generic image descriptor. Moreover, \cite{donahue2013decaf} showed that these features are so powerful that can be used for a variety of tasks in computer vision. Given an image as input we can apply a fully-convolutional neural network to obtain a feature vector per each receptive-field in the image\cite{long2015fully}. Since these features carry semantical information about the input image, they can be used to find image segments that are semantically coherent. In this paper, we show how these segments can be extracted from such CNN features.            

CNN features are very high-dimensional (namely, 4096). Traditional segmentation approaches that are mainly based on clustering techniques \cite{shi2000normalized} are not feasible. Since we want each segments corresponds to a meaningful visual element, large number of cluster centers are essential. That makes the segmentation process further more complex. To overcome such computational complexities, binary encoding of CNN features has proposed instead. A CNN feature is converted to a short binary code: each bit pattern represents a cluster center in the original CNN feature space. For example, an 32-bit binary code can generate $2^{32}$ clusters. Each bit corresponds to a visual attribute. Nearby pixels should have similar binary patterns unless they undergo a large semantical change. This is a perfect property to be used for semantic segmentation. Iterative-Quantization \cite{gong2013iterative} is employed to learn these binary codes. A powerful feature of the ITQ is that it generates bits in a simple way and the transformation is linear. This is perfect setting to be embedded in the CNN networks as a new layer. 
Once the binary map of the CNN features is available, a low-level superpixel extraction method is applied on the whole image and then the superpixels with the similar binary patterns (under Hamming distance) are merged together.


Our major contributions in this work can be summarized as; \emph{first:} We proposed a semantic segmentation which can be used in a general setting, unlike the all previous methods that are limited to specific categories. \emph{second:} We introduced a compact representation of high-dimensional CNN features in the form of binary codes, to preserve semantic information, thus can be used for semantic segmentation. Hence, we present a binary encoding layer in our network, which can updates using back-propagation. This new layer is able to be attached to any other deep-nets for encoding purposes.


\section{Related Work}
\label{sec:Related}

Despite a large body of works on low-level segmentation, there few works target semantic segmentation, and to the best of our knowledge, there is no work doing \emph{general} semantic segmentation utilizing high-level CNN features.

\noindent\textbf{Low-level segmentation:} 
Low-level segmentation refers to partitioning an input image into a set of perceptually meaningful atomic regions, considering the low-level image features, like intensity, edge, or texture. This step is usually considered as a pre-processing step which can effectively be employed to reduce complexity of subsequent visual recognition tasks. In literature, apart from the core low-level feature used, a substantial debate has been mainly posed over the optimization algorithms employed to efficiently solve this partitioning problem. In this context, two classes of approaches can be identified~\cite{achanta2012slic}.
On one hand, \emph{graph-based} methods treat pixels as nodes in graph, connected each other via edges reflecting their similarity in the feature space. Then, the graph is partitioned into a set of sub-graphs corresponding to image segments by minimizing a cost function. 
\begin{figure}[htb]
\centering
\centerline{\includegraphics[width=0.5\textwidth]{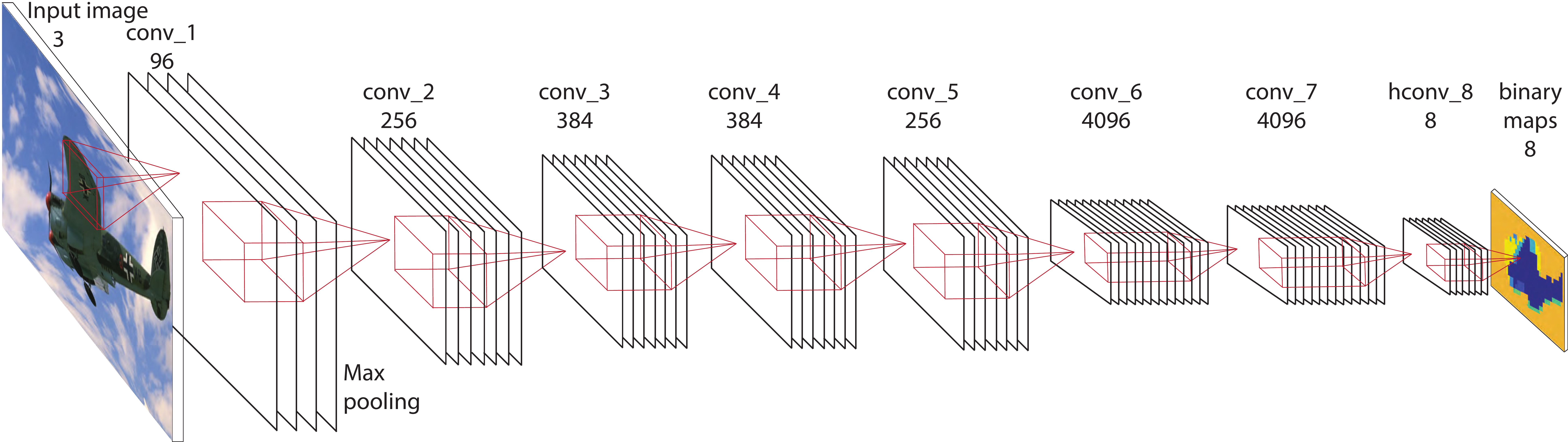}}
\caption{\small{Architecture of our proposed segmentation network.}}
\label{fig:net}
\end{figure}
Among the best performing methods, Normalized-Cuts~\cite{shi2000normalized}, Super-pixel Lattices~\cite{moore2008superpixel}, and Efficient Graph-based Segmentation (EGS)~\cite{felzenszwalb2004efficient} can be quoted. Here in this work, our method is compared with the last work, selected as one of the best performing non-semantic graph-based segmentation methods.

On the other hand, a different set of methods, named \emph{gradient-ascent-based} approaches, starts with an initial clusters of pixels, then refines iteratively until convergence in visual consistency. In this line of research, Mean-shift \cite{comaniciu2002mean}, Turbo-pixel \cite{levinshtein2009turbopixels} and state-of-the-art SLIC \cite{achanta2012slic} should be mentioned. Note that, SLIC has picked as one of our baselines and compare our method with.

\noindent\textbf{Semantic segmentation:} 
Recently, visual recognition task has come to rely increasingly on segmentation, and region extraction, accordingly, emerged to play a key role in object detection \cite{girshick2016region} and activity recognition\cite{jain2014action,gkioxari2015finding}. Due to the fact that the quality of initial segmentation affects significantly the subsequent tasks, providing segments capturing higher-level semantics became crucial. Semantic segmentation often formulated as combining low-level segments with region-based object detectors either in a cascade \cite{arbelaez2012semantic,carreira2012object,van2011segmentation} or joint \cite{dong2014towards,mottaghi2014role,yao2012describing} manner. Convolutional Neural Networks have recently resurfaced as a powerful tool for learning to segment semantically \cite{girshick2014rich,long2015fully,zheng2015conditional,chen2014semantic}. Nevertheless, learning such supervised deep structures for higher number of categories (and samples) is so supervision-demanding and computationally-expensive. Very recently, ADE20K~\cite{zhou2016semantic} has been introduced in which a wider variety of scenes and objects are annotated. Even in this case, extending the current supervised DNNs to work in a zero-shot fashion (namely, the categories other than the ones exist in the dataset) is not trivial. In this work, however, a completely different perspective to semantic segmentation is picked out. We specifically propose a method to narrow down the semantic gap (between pixels and concepts) in images, namely, trying to inject semantic inherited from generic CNN representations, so leading to more general semantic segmentation while maintaining the method complexity to a manageable level.

\section{CNN-aware binary map of image segments}
\label{sec:Method}
In this section, the two major parts of our proposed method are described in details: {\it 1) Spatial-aware fully convolutional network}, {\it 2) Binary map encoding layer} and {\it 3) Semantic segmentation using binary maps}. Figure~\ref{fig:overview} illustrates a general work-flow of the method. 

\subsection{Spatial-aware fully convolutional network}
\label{sec:convNet}
Early convolutional layers in CNNs represent more local information of the image, while deeper ones contain more global information. The fully-connected layers capture higher-level information and usually employed for recognition purposes. It has been shown that the deep nets which trained on ImageNet~\cite{deng2009imagenet}, are rather semantic; they can address wide range of recognition problems~\cite{razavian2014cnn,donahue2013decaf}. Fully convolutional Nets also can preserve relative spatial coordinates between input image and output feature map. These properties motivated us to use such structures for general semantic segmentation.


For the sake of generalization, we adopted a pre-trained network (Namely, AlexNet~\cite{alexnet}) and simply converted it to a fully convolutional net. It provides us with general, yet spatially-consistent semantic representations. Due to the semantic power of the \emph{fc7} layer in case of AlexNet, we exploited the corresponding layer in our network (denoted by $conv7$) to extract feature maps.
\subsection{Binary map encoding layer}
\label{sec:binary}

Clustering the extracted high-dimensional feature maps from $conv7$ comes with high computational cost. It leads to converge to a limited number of clusters. One possible solution to avoid this problem is partitioning high-dimensional features into a set of buckets (instead of clusters) using hashing techniques. It provide use with generating small binary codes for each feature vector, taking into account their distance simultaneously. Assuming 24-bits of binary code can address $2^{24}$ buckets. Moreover, this binary map can be represented as a 3-channels RGB image, providing a better illustration on partitions. Obviously, dealing with binary codes comes with lower computational cost and higher efficiency with respect to other clustering methods. However, the most advantage of hashing comparing to clustering, is the capability of embedding it simply as a layer inside the network.\\
\noindent\textbf{Binary encoding layer:}
Encoding feature maps to binary codes is computed by Iterative Quantization Hashing (ITQ)\cite{itq}. It is a unsupervised binary codes method, which projects each high-dimensional feature vector into a binary space. The last layer of our network (denoted by $hconv8$) is built by the hashing linear transformation, learned initially by ITQ. During testing, for each input image to the network, a spatial-aware binary map would be generated. For this purpose, we forward-pass the image through all the convolutional layers as well as the final binary encoding layer. Such binary map is eventually used for the segmentation purpose.

\begin{figure}
\subfigure[Image Samples] {
\includegraphics[scale=0.18]{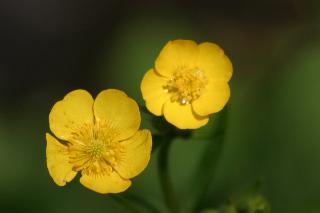}
\includegraphics[scale=0.18]{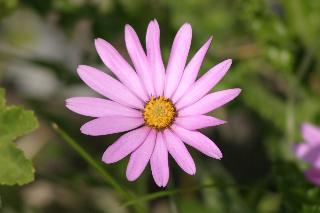}
\includegraphics[scale=0.16]{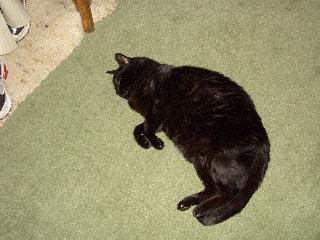}
\includegraphics[scale=0.103]{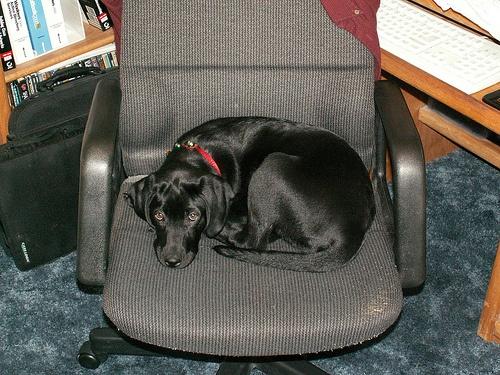}
}
\subfigure[Ground truth] {
\includegraphics[scale=0.18]{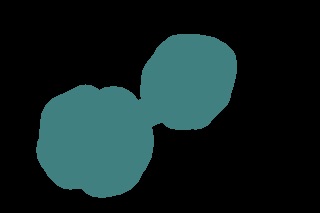}
\includegraphics[scale=0.18]{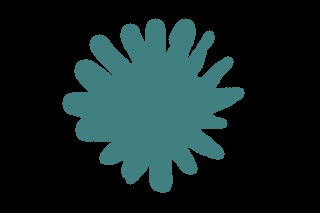}
\includegraphics[scale=0.16]{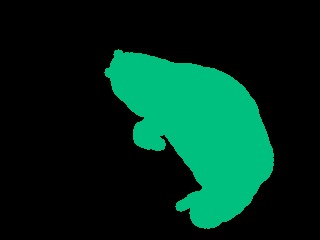}
\includegraphics[scale=0.103]{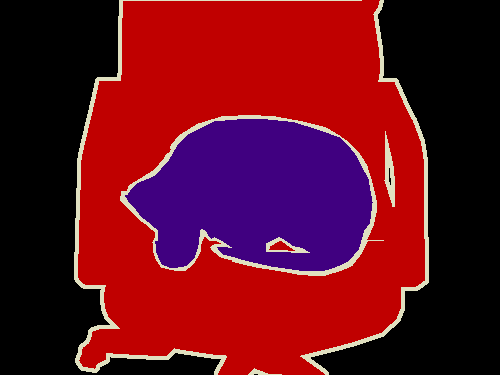}
}
\subfigure[Efficient Graph-based Segmentation (EGS)] {
\includegraphics[scale=0.18]{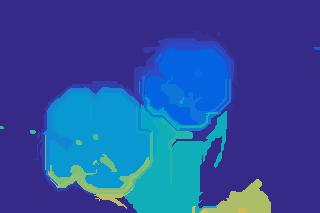}
\includegraphics[scale=0.18]{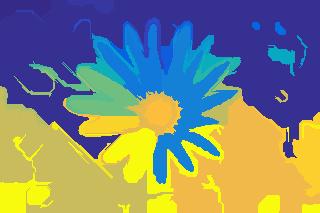}
\includegraphics[scale=0.16]{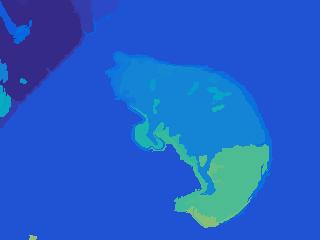}
\includegraphics[scale=0.103]{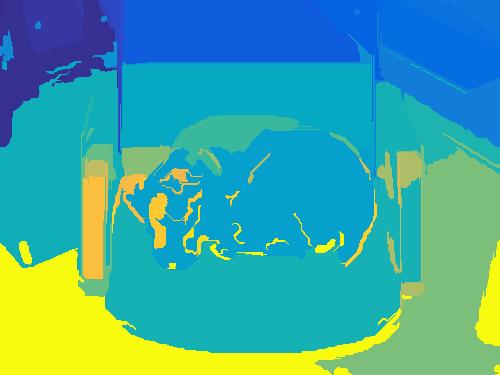}
}
\subfigure[Binary maps visualization] {
\includegraphics[width=2.04cm,height=1.5cm]{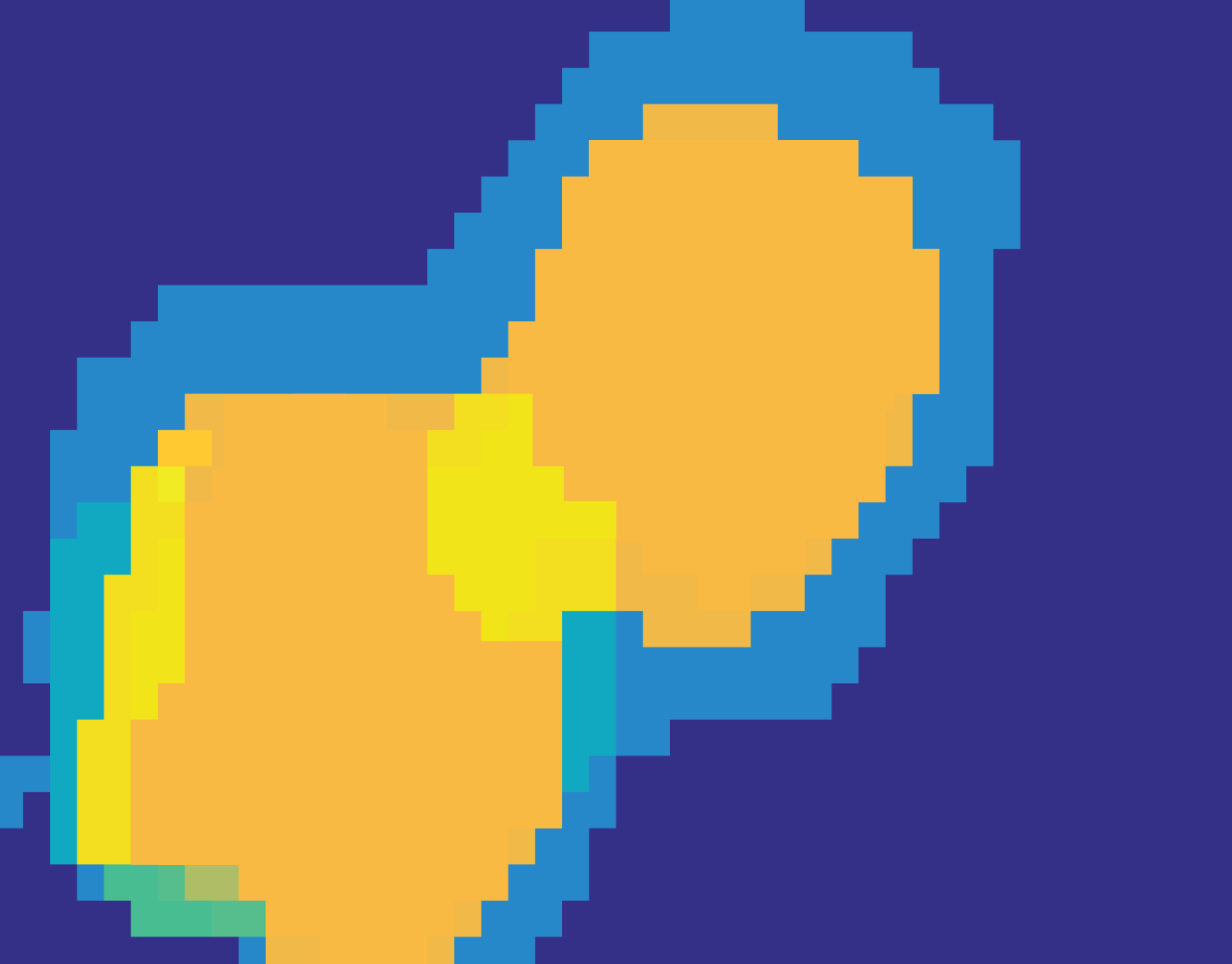}
\includegraphics[width=2.04cm,height=1.5cm]{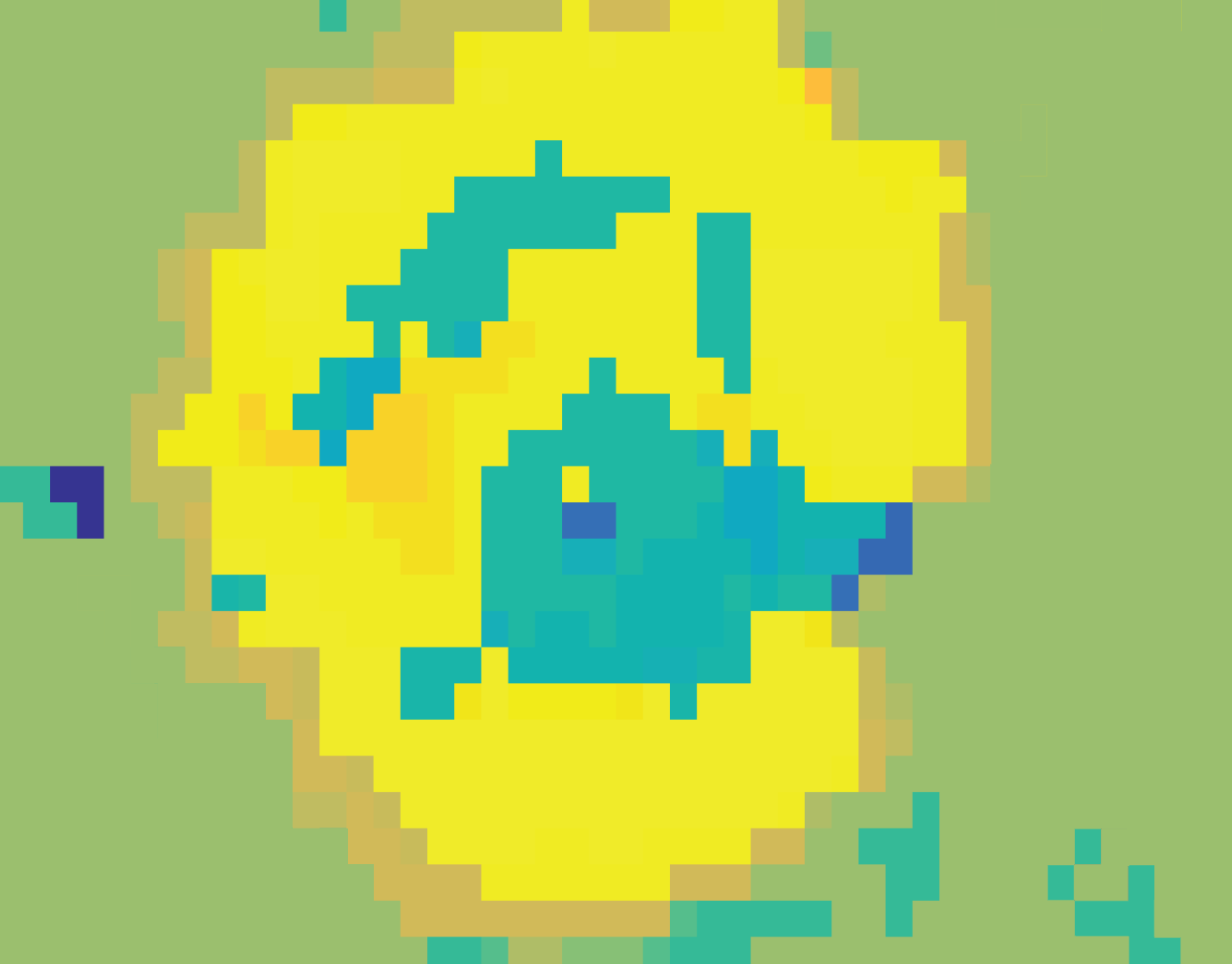}
\includegraphics[width=1.8cm,height=1.5cm]{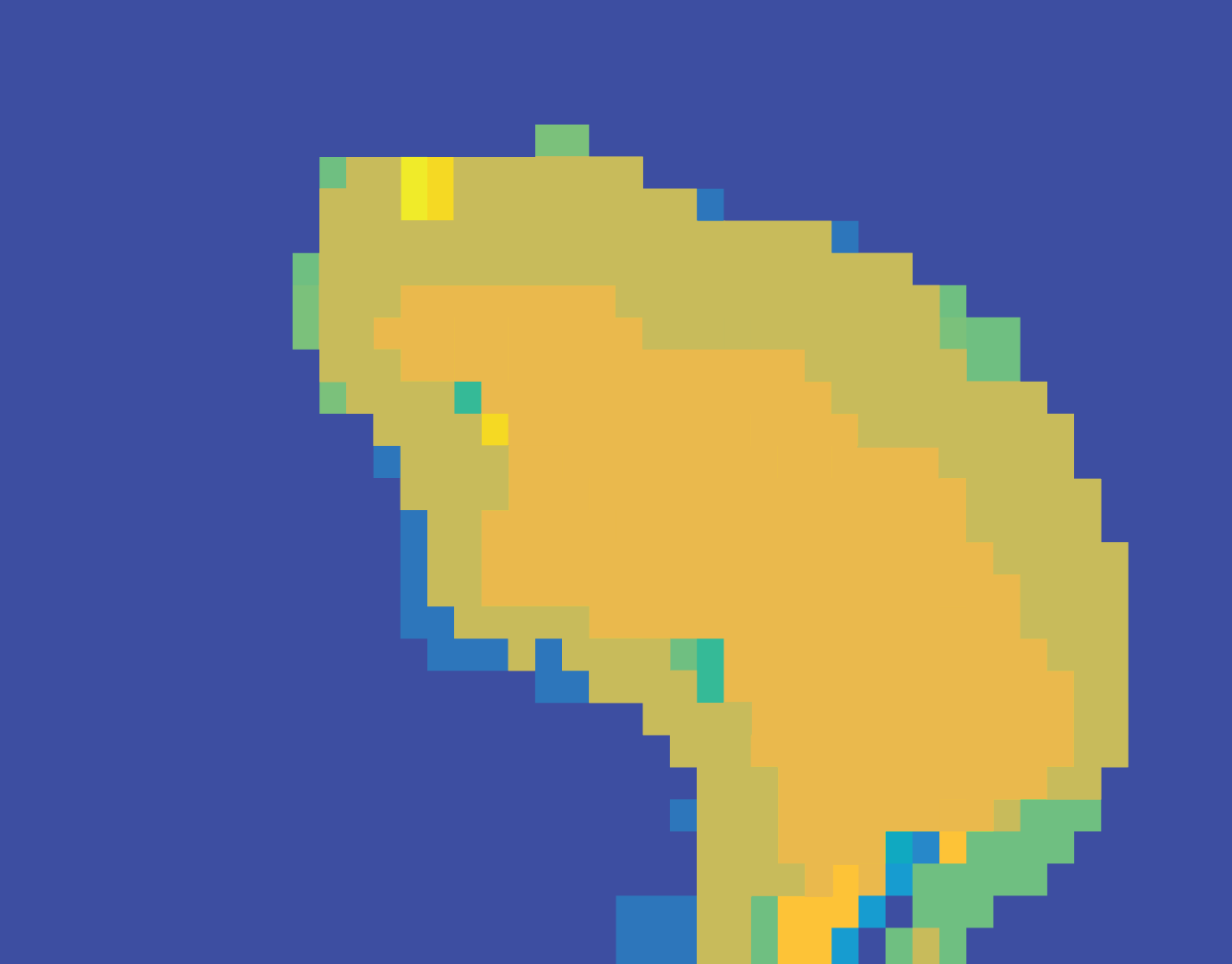}
\includegraphics[width=1.82cm,height=1.5cm]{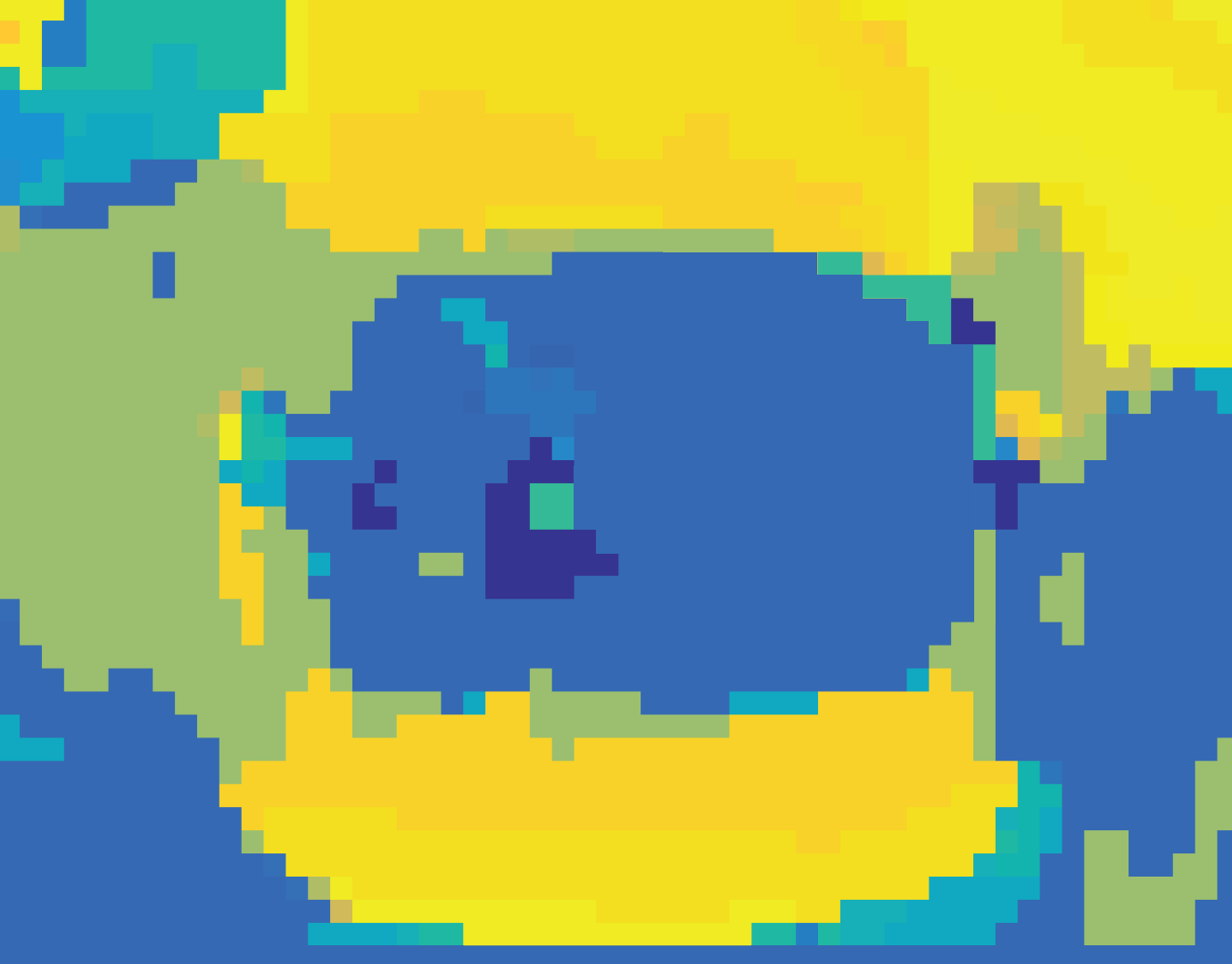}
}
\subfigure[Our method] {
\includegraphics[scale=0.18]{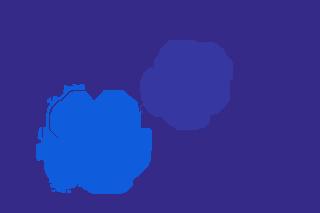}
\includegraphics[scale=0.18]{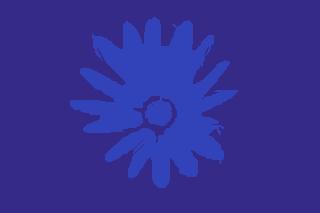}
\includegraphics[scale=0.16]{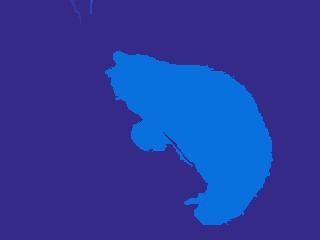}
\includegraphics[scale=0.103]{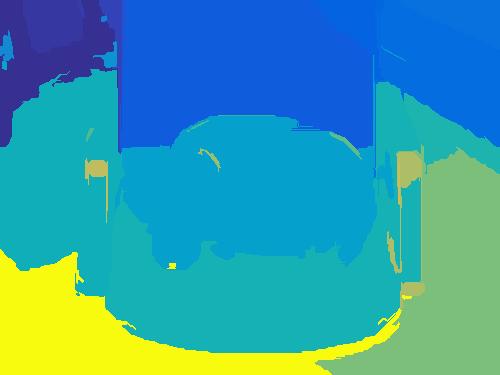}
}
\caption{\small{Our segmentation method compare with EGS \cite{felzenszwalb2004efficient}}}
\label{fig:db}    
\end{figure}
\subsection{Semantic segmentation using binary map}
\label{sec:segmentation}
The generated binary maps have two important aspects: {\it first;} it preserves the spatial relation between input image and output features. In other word, each region on the binary map corresponds to a patch on the input image. {\it second;} binary maps are generated using the convolutional feature maps of the deep-net, hence they capture the semantics of the scene. Binary pattern with different values represent areas with different semantics. It specifically interesting because, any changes in the binary code patterns on binary maps can be interpreted as a semantics change on the corresponding areas on the image.\\
We take advantage of these two properties to employ the binary map for semantic segmentation. To this purpose, we initialized the segmentation by low-level superpixels, then merged superpixels with the similar binary codes in the binary map. This simple yet  effective criteria on semantic features has shown to be much more powerful compares to the previous state-of-the-art methods which utilized the sophisticated partitioning algorithms but relaying only on low-level visual information.



\section{Experiment}
\label{sec:Experiment}
A set of experiments has designed to evaluate our method. This section, explains the experimental setup, evaluation protocol, datasets, and finally elaborate on the results.

\noindent\textbf{Experimental Setup:}
Set of comparative experiments has been done to demonstrate the advantage of using our binary map for semantic segmentation compares to two best-performing low-level segmentation methods: \emph{Efficient Graph-Based Segmentation (EGS)}\cite{felzenszwalb2004efficient} and \emph{gradient-ascent-based SLIC}\cite{achanta2012slic}.
Since there is no pre-processing step or parameter setting in our method, 
For the baselines, we used the publicly available codes with the default parameters ($\sigma = 1.0, k = 100$). In this means, we aims at showing the strength of the semantic segmentation compares to low-level segmentation without any parameter tuning.

\noindent\textbf{Dataset:}
We choose two datasets; The first dataset is the Berkeley Segmentation Dataset (BSDS500), includes 300 training samples as and 200 images for testing. The other, Microsoft Research Cambridge database (MSRC), includes 510 images. Both evaluation has been performed with the original setups of datasets.

\noindent\textbf{Evaluation Protocol:} we adopted the Segmentation Intersection over Union (IoU) as one of the most commonly used evaluation measure for segmentation task. IoU describe as:
\begin{center}
$IoU(P_{m},P_{gt}) = \frac{|P_{m} \bigcap P_{gt}|}{|P_{m} \bigcup P_{gt}|}$
\end{center}
Where $P_{gt}$ is ground truth segment annotation, and $P_{m}$ is predicted segment. As the predicted segments, we select the segments with the maximum IoU with each segment in $P_{gt}$. The final value of \emph{Segmentation-IoU} is computed as the average over all the segments of all the images of the dataset.

\noindent\textbf{Segmentation Network Details:}
The designed deep-net consist of two major parts; \emph{1) Fully convolutional network:} 
As reviewed in~\ref{sec:convNet} at first we utilized a pre-trained AlexNet model on ImageNet. Original AlexNet, contains 5 convolutional layers and two fully connected layers. In order to obtain spatial-aware feature maps, we convert the last two fully connected layers into convolutional layers. By transforming fully connected layers into convolutional layers we could enable the net to output a multi dimensional feature map disregard to input image size and produce an efficient model for spatial-aware patch pooling. In our experiment we used images with higher size to produce finer feature maps. High dimensional feature maps(28x44x4096) extracted from the last convolutional layer feed to a new layer which we called {\it Binary bit-map Layer} to compress into a lower dimension binary bit map (28x44x8).
\emph{2) Binary Bit-map layer:}
Bit-map layer is designed for convolutional feature map quantization. In order to build the layer, we first extract convolutional maps from $conv7$ over PASCAL 2007 images to train an unsupervised ITQ hash to model 4096 dimensional feature maps into 8-bits binary codes. Hashing weights obtain from ITQ applied into a Depth Normalization Layer. We embedded the Depth Normalization Layer with pre-trained weights to the network to build {\it Binary Bit-map layer}. Output of the network is a set of 8-bit binary maps. Figure~\ref{fig:overview} shown a visualization example of extracted binary map in the form of grayscale image, which is spacial aware and contains semantic information about the image. This bit-mat image is eventually utilized for image segmentation.
\begin{figure}
\subfigure[Segmentation-IoU on Berkeley] {
\begin{tikzpicture}[thick,scale=0.44]
\begin{axis}[
xlabel={Number of Super-pixels},
ylabel={Segmentation accuracy \%},
xmin=90, xmax=510,
ymin=30, ymax=60,
xtick={100,200,300,400,500},
ytick={30,35,40,45,50,55,60},
legend style={at={(1,1)},anchor=north east},
ymajorgrids=true,
grid style=dashed]
\addplot+[
color=blue,
mark=square,
]
coordinates {
	(100,38.7)(200,48.35)(300,42.5)(400,44.78)(500,44.35)
};
\addplot+[
color=red,
mark=square,
]
coordinates {
	(100,34.1)(200,45.19)(300,40.4)(400,44.78)(500,42.35)
};
\addplot+[
color=green,
mark=square,
]
coordinates {
	(100,37.8)(200,41.7)(300,41.9)(400,43.7)(500,39.12)
};
\legend{Our Method, EGS Method, Slice Method}
\end{axis}
\end{tikzpicture}
}
\subfigure[Segmentation-IoU on MSRC] {
\begin{tikzpicture}[thick,scale=0.44]
\begin{axis}[
xlabel={Number of Super-pixels},
ylabel={Segmentation accuracy \%},
xmin=90, xmax=510,
ymin=30, ymax=65,
xtick={100,200,300,400,500},
ytick={30,35,40,45,50,55,60},
legend style={at={(1,1)},anchor=north east},
ymajorgrids=true,
grid style=dashed]
\addplot+[
color=blue,
mark=square,
]
coordinates {
	(100,44.5)(200,52.83)(300,55.03)(400,53.17)(500,52.1)
};
\addplot+[
color=red,
mark=square,
]
coordinates {
	(100,39.23)(200,47.12)(300,48.3)(400,47.3)(500,46)
};
\addplot+[
color=green,
mark=square,
]
coordinates {
	(100,37.77)(200,42.3)(300,45.23)(400,45.1)(500,44.9)
};
\legend{Our Method, EGS Method, Slice Method}
\end{axis}
\end{tikzpicture}
}
\caption{\small{Segmentation-IoU over superpixel variation}}
\label{tab:segacc}
\end{figure}
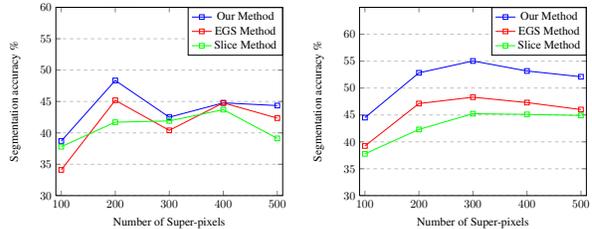

\noindent\textbf{Segmentation Strategy:}
For segmentation, we first extract the binary maps for each input image. For each superpixel a binary code is assigned by the corresponding region on binary map. Then we merged the superpixels with the similar binary codes on the bit maps (i.e., zero distance in Hamming space), Figure \ref{fig:db}(c). The final segmentation is obtained as the result of such merging of superpixels, Figure \ref{fig:db}(e).\\
\noindent\textbf{Experimental Result:}
Our proposed semantic segmentation significantly outperform previous low-level segmentation methods, both quantitatively, and qualitatively. A comparison results on two datasets demonstrated in Table \ref{tab:compare}. The first column shows a comparison of average segmentation-IoU for the algorithms in the MSRC dataset, and the second columns compares the algorithms on BSDS500. Qualitative comparison with EGS method, Figure \ref{fig:db}, also shows the better results achieved by our approach. Figure \ref{fig:db} visualize the results of our method and EGS method. We outperform both baseline methods by large margins in term of $segmentation-IoU$ over different superpixel sizes (Figure \ref{tab:segacc}). Such evaluation shows the robustness of the proposed method to the number of super-pixels.We observe that segmentation on images containing ``things'' (objects) are significantly better as compared to images containing ``stuffs'' (scenes). It also supports our hypothesis that binary patterns preserved semantic information and the understanding objects in the scene.

\begin{table}
	\begin{center}
		\begin{tabular}{|cc|cc|}
			\hline
			\multicolumn{2}{|c|}{\texttt{MSRC}} & \multicolumn{2}{c|}{\texttt{Berkeley}} \\
			\hline
			\hline
			Method & IoU $\;\;$& Method & IoU $\;\;$ \\
			\hline
			EGS \cite{felzenszwalb2004efficient}  	& 50.3\% 	& EGS \cite{felzenszwalb2004efficient} & 45.19\% \\
			SLIC  \cite{achanta2012slic} 	& 48.7\% 	& SLIC  \cite{achanta2012slic} & 43.70\% \\
			\hline
			\hline
			Our method& \textbf{55.03 \% }& Our method & \textbf{48.35\%}  \\
			\hline
			
		\end{tabular}
	\end{center}
	\caption{\small{Quantitative results on MSRC and Berkeley datasets.}}
	\label{tab:compare}
\end{table}


\section{Conclusion}
\label{sec:conclusion}
In this work a novel approach to general semantic-aware image segmentation has been presented which does no require category-specific training a deep-net. We employed AlexNet as pre-trained model and convert fully connected layers into convolutional layers. An efficient ITQ hashing layer is attached as the final layer to the net to quantity high dimensional feature maps in form of binary code representation. Such model provides both spatial consistency as well as low dimensional semantic embedding. Our experimental results shown using these binary maps can improve the performance of the segmentation comparing to several low-level segmentation methods. As future work, we will study fine tuning the hashing layer with back-propagation and end-to-end training of semantic segmentation net with recent Region Proposal Network (RPN) in a joint manner.

\bibliographystyle{IEEEbib}
\bibliography{strings,refs}

\end{document}